\documentclass[preprint,12pt]{elsarticle}

\usepackage{amssymb}
\usepackage{amsthm}
\usepackage{url}
\usepackage{indentfirst}
\usepackage{url}

\newtheorem{theorem}{Theorem}

\newtheorem{corollary}{Corollary}
\newtheorem{lemma}{Lemma}

\def\argmin{{\rm argmin}}

\def\argmin[{{\rm argmin}}

\def\x{{\bf x}}
\def\y{{\bf y}}

\def\p{{\bf p}}
\def\q{{\bf q}}

\def\w{{\bf w}}
\def\l{{\bf l}}
\def\0{{\bf 0}}
\def\c{{\bf c}}

\def\c{{\bf c}}

\def\F{{\bf F}}
\def\subst{{\rm Subst}}

\def\0{{\bf 0}}

\journal{Neurocomputing}

\begin{document}

\begin{frontmatter}

\title{Long-Term Online Smoothing Prediction\\Using Expert Advice}
\author{Alexander Korotin}
\ead{a.korotin@skoltech.ru}
\author{Vladimir V'yugin}
\ead{v.vuygin@skoltech.ru}
\author{Evgeny Burnaev}
\ead{e.burnaev@skoltech.ru}
\address{Skolkovo Institute of Science and Technology}

\begin{abstract}
 For the prediction with experts' advice setting we construct forecasting algorithms that suffer loss not much more than any expert in the pool. In contrast to the standard approach, we investigate the case of long-term forecasting of time series and consider two scenarios. In the first one, at each step $t$ the learner has to combine the point forecasts of the experts issued for the time interval $[t+1, t+d]$ ahead. Our approach implies that at each time step experts issue point forecasts for arbitrary many steps ahead and then the learner (algorithm) combines these forecasts and the forecasts made earlier into one vector forecast for steps $[t+1,t+d]$. By combining past and the current long-term forecasts we obtain a smoothing mechanism that protects our algorithm from temporary trend changes, noise and outliers. In the second scenario, at each step $t$ experts issue a prediction function, and the learner has to combine these functions into the single one, which will be used for long-term time-series prediction. For each scenario we develop an algorithm for combining experts forecasts and prove $O(\ln T)$ adversarial regret upper bound for both algorithms.
 
\end{abstract}


\end{frontmatter}


\section{Introduction}

The problem of long-term forecasting of time series is of high practical importance. For example, nowadays nearly everybody uses long-term weather forecasts \cite{richardson2007weather, lorenc1986analysis} (24 hour, 7 days, etc.) provided by local weather forecasting platforms. Road traffic and jams forecasts 
\cite{de2015grenoble, herrera2010evaluation,myr2002real}
are being actively used in many modern navigating systems. Forecasts of energy consumption and costs \cite{gaillard2015forecasting}, 
web traffic \cite{oliveira2016computer} and stock prices
\cite{ding2015deep,pai2005hybrid} are also widely used in practice.




Many state-of-the-art (e.g. ARIMA \cite{box2015time}) and modern (e.g. Facebook Prophet\footnote{\url{https://github.com/facebook/prophet}} \cite{taylor2018forecasting}) time series forecasting approaches produce a model 
that is capable of predicting arbitrarily many steps ahead. The advantage of such 
models is that when building the final forecast at each step $t$ for interval 
${[t+1, t+d]}$ ahead, one may use forecasts made earlier at the steps ${\tau<t}$. Forecasts of each step ${\tau<t}$ are made using 
less of the observed data. Nevertheless, they can be more robust to noise, outliers 
and novelty of the time interval ${[\tau+1, t]}$. Thus, the usage of such outdated 
forecasts may prove useful, especially if time series is stationary.

In general, we consider the game-theoretic on-line learning model in which a master (aggregating) 
algorithm has to combine predictions from a set of experts. The problem setting we 
investigate can be considered as the part of Decision-Theoretic Online Learning (DTOL)
or Prediction with Expert Advice (PEA) framework (see e.g. \cite{LiW94,FrS97,Vov90,VoV98,cesa-bianchi,korotin2018aggregating} 
among others). In this framework the learner is usually called the aggregating algorithm. 
The aggregating algorithm combines the predictions from a set of experts in the online 
mode during time steps $t=1,2,\dots,T$. 

In practice for time series prediction the square loss function is widely used. The square loss function is mixable \cite{VoV98}. 
For mixable loss functions Vovk's  aggregating algorithm (AA) \cite{VoV98, VoV2001} is the most appropriate, 
since it has theoretically best performance among all known algorithms. We use the aggregating algorithm
as the base and modify it for the long-term forecasting.

The long-term forecasting considered in this paper is a case of the forecasting with a delayed feedback. 
As far as we know, the problem of the delayed feedback forecasting was first considered 
by \cite{WeO2002}. 


In this paper we consider the two scenarios of the long-term forecasting. In the first one, at each step $t$ the learner has to combine the point forecasts of the experts issued for the time interval $[t+1, t+d]$ ahead. In the second scenario, at each step $t$ experts issue prediction functions, and the learner has to combine these functions into the single one, that will be used for long-term time-series prediction.

The first theoretical problem we investigate in the paper is the effective usage of the outdated forecasts. Formally, the learner is given $N$ basic 
forecasting models. Each model ${n=1,2,\dots,N}$ at every step $t$ produces infinite 
forecast for the steps ${t+1, t+2,\dots}$ ahead.
The goal of the learner at each step $t$ is to combine the current models' forecasts 
and the forecasts made earlier into one aggregated long-term forecast for the time 
interval $[t+1, t+d]$ ahead. We develop an algorithm to efficiently combine these forecasts.


Our main idea is to replicate any expert $n$ in an infinite sequence of auxiliary experts 
$(n,\tau)$, where $\tau=1,2,\dots$. Each expert $(n,\tau)$ issues at time moment $\tau$ 
an infinite sequence of forecasts for time moments $\tau+1,\tau+2,\dots$. 
Only a finite number of the experts are available at any time moment. The setting 
presented in this paper is valid also in case where only one expert ($N=1$) is given. 
At any time moment $t$ the AA uses predictions of each expert $(n,\tau)$ for the time 
interval $[t+1, t+d]$ (made by expert $n$ at time 
$\tau\leq t$). In our case, the performance of the AA on the step $t$ is measured by 
the regret $r_{t}$ which is the difference between the average loss of the aggregating algorithm 
suffered on time interval $[d+1,T]$ and the average loss of the best auxiliary expert $(n,\tau)$ 
suffered on the same time interval. 
Note that the recent related work is \cite{Kaln2017} where an 
algorithm with tight upper bound for predicting vector valued outcomes was presented.

In the second part of our paper we consider the online supervised learning scenario. The data is represented by pairs $(x,y)$ of predictor-response variables. Instead of point or interval predictions, 
the experts and the learner present predictions in the form of functions $F(x)$ from signals $x$. Signals $x_t$ appear gradually over time $t$ and allow to calculate forecasts as the values $F(x_t)$ of these functions.
For this problem we present method for smoothing regression using expert advice. 

The article is structured as follows. In Section~\ref{prel-1} we give some preliminary notions. 
In Section~\ref{v-s-1} we present the algorithm for combining long-term forecasts of 
the experts. Theorem~\ref{main-3fa} presents a performance bound $O(\log (NT))$ for the regret 
of the corresponding algorithms. 

In Section~\ref{regr-1} we apply PEA approach for a case of the online supervised learning and develop an algorithm for online smoothing regression. Also, we provide experiments conducted on synthetic data and show the effectiveness of the proposed method.
In \ref{vector-pred-1} some auxiliary results are presented.

\section{Preliminaries}\label{prel-1}

In this section we recall the main ideas of prediction with expert advice theory.
Let a pool of $N$ experts be given.
Suppose that elements $y_1,y_2,\dots$ of a time series are revealed 
online -- step by step.
Learning proceeds in trials $t = 1,\ldots, T$. At each time moment $t$ experts 
$i\in\{1,\dots ,N\}$ present their predictions $c^i_t$ and the aggregating algorithm 
presents its own forecast $\gamma_t$. When the corresponding outcome(s) are revealed, 
all the experts suffer their losses using a loss function: 
$l^i_t=\lambda(y_t,c^i_t)$, $i=1,\dots ,N$. Let $h_t=\lambda(y_t,\gamma_t)$ 
be the loss of the aggregating algorithm. The cumulative loss suffered by any expert $i$ 
and by AA during $T$ steps are defined as
\begin{eqnarray*}
L_T^i = \sum\limits_{t=1}^T l_t^i\mbox{   and   } H_T = \sum\limits_{t=1}^T h_t.
\end{eqnarray*}
The performance of the algorithm w.r.t. an expert $i$ can be measured by the regret 
$R^i_T=H_T-L^i_T$.


The goal of the aggregating algorithm is to minimize the regret with respect to
each expert. In order to achieve 
this goal, at each time moment $t$, the aggregating algorithm evaluates performance 
of the experts in the form of a vector of experts' weights 
$\w_t=(w_{1,t},\dots ,w_{N,t})$, where $\|\w_{t}\|_{1}=1$ and $w_{i,t}\ge 0$ for all $i$.
The weight $w_{i,t}$ of an expert $i$ is an estimate of the quality of the expert's 
predictions at step $t$. In classical setting (see \cite{FrS97}, \cite{Vov90} 
among others), the process of expert $i$ weights updating is based on the method of 
exponential weighting with a learning rate $\eta>0$:
\begin{eqnarray}
w^\mu_{i,t}=\frac{w_{i,t}e^{-\eta l_t^i}}{\sum\limits_{j=1}^N w_{j,t}e^{-\eta l_t^j}},
\label{weight-update-1}
\end{eqnarray}
where $\w_1$ is some weight vector, for example, ${\w_1=(\frac{1}{N},\dots ,\frac{1}{N})}$.
In classical setting, we prepare weights $\w_{t+1}=\w^\mu_t$ for using at the next step
or, in a more general case of the $d$-th outcome ahead 
prediction, we define ${\w_{t+d}=\w^\mu_t}$, where $d\ge 1$.


The Vovk's aggregating algorithm (AA) (\cite{Vov90}, \cite{VoV98}) is the base 
algorithm in our study. Let us explain the main ideas of learning with AA.

We consider the learning with a mixable loss function $\lambda(y, \gamma)$. 
Here $y$ is an element of some set of 
outcomes $y$, and $\gamma$ is an element of some set of forecasts $\Gamma$. 
The experts $1\le i\le N$ present the forecasts $c_i\in\Gamma$. 

In this case the main tool is a superprediction function
\begin{eqnarray*}
g(y)=-\frac{1}{\eta}\ln\sum\limits_{i=1}^N e^{-\eta\lambda(y,c_i)}p_i,
\end{eqnarray*}
where $\p=(p_1,\dots ,p_N)$ is a probability distribution on the set of all experts and 
$\c=(c_1,\dots , c_N)$ is a vector of the experts predictions.

The loss function $\lambda$ is mixable if for any probability distribution $\p$ on 
the set of experts and for any set of experts predictions $\c$ a value of $\gamma$ exists such that 
\begin{eqnarray}
\lambda(y,\gamma)\le g(y) 
\label{subst-1}
\end{eqnarray}
for all $y$.

We fix some rule $\gamma=\subst(\c,\p)$ for computing a forecast
satisfying (\ref{subst-1}). $\subst$ is called a substitution function.

It will be proved in Section \ref{vector-pred-1} that using the rules 
(\ref{weight-update-1}) and (\ref{subst-1}) for defining weights and the forecasts
in the online mode we obtain 
$$
H_T\le\min_{1\le i\le N} L^i_T+\frac{\ln N}{\eta}
$$
for all $T$.

A loss function $\lambda(y,\gamma)$ is $\eta$-exponential concave if for 
any $y$ the function $e^{-\eta\lambda(y,\gamma)}$ is concave w.r.t. 
$\gamma$. By definition any $\eta$-exponential concave function is $\eta$-mixable.

The square loss function $\lambda(y,\gamma)=(y-\gamma)^2$ is $\eta$-mixable
for any $\eta$ such that $0<\eta\le\frac{1}{2B^2}$, where $y$ and $\gamma$ 
are a real numbers and $y\in [-B,B]$ for some $B>0$, 
see \cite{Vov90,VoV98}.

By \cite{VoV98} and \cite{VoV2001}, for the square loss function,  
the corresponding forecast can be defined as
\begin{eqnarray}
\gamma=\subst(\c,\p)=\frac{1}{4B}(g(-B)-g(B))=
\frac{1}{4\eta B}\ln\frac{\sum\limits_{i=1}^N p_i e^{-\eta (B-c_i)^2}}
{\sum\limits_{i=1}^N p_i e^{-\eta (B+c_i)^2}}.
\label{subst-1a}
\end{eqnarray}
For the $\eta$-exponential concave loss function we can also use a more straightforward 
expression for the substitution function:
\begin{eqnarray}
\gamma=\subst(\c,\p)=\sum_{i=1}^N c_i p_i. 
\label{subst-2a}
\end{eqnarray}
The inequality (\ref{subst-1}) also holds for all $y$. 

The square loss function is $\eta$-exponential concave for $0<\eta\le\frac{1}{8B^2}$.
However, the definition (\ref{subst-2a}) results in four times more regret
(see \cite{KiW99} and Section \ref{vector-pred-1}).

\section{Algorithm for Combining Long-term Forecasts of Experts.}\label{v-s-1}

In this section we consider an extended setting. At each time moment $t$ each 
expert $n\in\{1,\dots ,N\}$ presents an infinite sequence 
$\c^n_t=(c^n_{t,1},c^n_{t,2},\dots)$ of forecasts for the time moments 
$t+1,t+2,t+3,\dots$. A sequence of the corresponding confidence levels 
$\p^n_t=(p^n_{t,1},p^n_{t,2},\dots)$ also can be presented at time moment $t$. 
Each element of this sequence is a number between 0 and 1. 
If $p^n_{t,i}<1$, then it means that we use the forecast $p^n_{t,i}$ only partially 
(e.g. it may become obsolete with time). If  $p^n_{t,i}=0$ then the corresponding 
forecast is not taken into account at all.\footnote{For example, in applications, 
it is convenient for some $d$ to set $p^n_{t,i}=0$ for all $i>d$, since too far 
predictions become obsolete.}
Confidence levels can be set by the expert itself or by the learner.\footnote{
The setting of prediction with experts that report their confidences as a number
in the interval $[0,1]$ was first studied by \cite{BlM2007} and further developed
by \cite{CBMS2007}.}

At each time moment $t$ we observe sequences $\c^n_{\tau}$, $\p^n_{\tau}$ issued by 
the experts $1\le n\le N$ at the time moments $\tau\le t$. To aggregate the forecasts 
of all experts, we convert any ``real'' expert $n$ into the infinite sequence of 
the auxiliary experts $(n,\tau)$, where $1\le\tau<\infty$.

At each time moment $t$ expert $(n,\tau)$ presents his forecast which is the segment 
of the sequence $\c^n_\tau$ of length $d$ starting at its $(t-\tau+1)$th element. 
More precisely, the forecast of the auxiliary expert $(n,\tau)$ is a vector
$$
\c^{(n,\tau)}_t=(c^{(n,\tau)}_{t,1},\dots,c^{(n,\tau)}_{t,d}),
$$
where for ${1\le s\le d}$ we set $c^{(n,\tau)}_{t,s}=c^n_{\tau,t-\tau+s}$.

We also denote the corresponding segments of confidence levels by 
$$
{\p^{(n,\tau)}_t=(p^{(n,\tau)}_{t,1},\dots,p^{(n,\tau)}_{t,d})},
$$
where $p^{(n,\tau)}_{t,s}=p^n_{\tau,t-\tau+s}$ for $1\le\tau\le t$ and 
$p^{(n,\tau)}_{t,s}=0$ for $\tau>t$.

Using the losses suffered by the experts $(n,\tau)$ (for ${\tau\le t}$)
on the time interval $[t-d+1,t]$, the aggregating algorithm updates the weights 
$w_{(n,\tau),t}$ of all the experts $(n,\tau)$ by the rule (\ref{weight-update-1}).
We denote these weights by $w_{(n,\tau),t+d}$ and use them for computing the aggregated
interval forecast for $d$ time moments $t+1,\dots ,t+d$ ahead
$$
\gamma_t=(\gamma_{t,1},\dots ,\gamma_{t,d}).
$$ 
We use the fixed point method by \cite{ChV2009}. Define the virtual forecasts of 
the experts $(n,\tau)$:
\[
\tilde c^{(n,\tau)}_{t,s}=
\left\{
    \begin{array}{l}
      c^{(n,\tau)}_{t,s}\mbox{ with probability } p^{(n,\tau)}_{t,s},
    \\
      \gamma_{t,s} \mbox{ with probability } 1-p^{(n,\tau)}_{t,s}.
    \end{array}
  \right.
\]
where $1\le n\le N$ and $1\le\tau<\infty$.

We consider any confidence level $p^{(n,\tau)}_{t,s}$ as a probability distribution
$\p^{(n,\tau)}_{t,s}=(p^{(n,\tau)}_{t,s}, 1-p^{(n,\tau)}_{t,s})$ on a two element
set.

First, we provide a justification of the algorithm presented below.
Our goal is to define the forecast $\gamma_t=(\gamma_{t,1},\dots,\gamma_{t,d})$
such that for $s=1,\dots ,d$ 
\begin{eqnarray}
e^{-\eta\lambda(y,\gamma_{t,s})}\ge\sum_{n=1}^N\sum_{\tau=1}^\infty 
E_{\p^{(n,\tau)}_{t,s}}[e^{-\eta\lambda(y,\tilde c^{(n,\tau)}_{t,s})}]
w_{(n,\tau),t+d}
\label{for-1b-1}
\end{eqnarray}
for each outcome $y$. Here $E_{\p_{(n,\tau),s}}$ is the mathematical expectation 
with respect to the probability distribution $\p_{(n,\tau),s}$. Also, 
$w_{(n,\tau),t+d}$ is the weight of the auxiliary expert $(n,\tau)$
accumulated at the end of step $t$.

We rewrite inequality (\ref{for-1b-1}) in a more detailed form: for any $1\le s\le d$,
\begin{eqnarray}
e^{-\eta\lambda(y,\gamma_{t,s})}\ge
\nonumber
\\
\sum_{n=1}^N\sum_{\tau=1}^\infty 
E_{\p^{(n,\tau)}_{t,s}}[e^{-\eta\lambda(y,\tilde c^{(n,\tau)}_{t,s})}]
w_{(n,\tau),t+d}=
\label{cond-1}
\\
\sum_{n=1}^N\sum_{\tau=1}^t p^{(n,\tau)}_{t,s}w_{(n,\tau),t+d}
e^{-\eta\lambda(y,c^{(n,\tau)}_{t,s})}+
\nonumber
\\
e^{-\eta\lambda(y,\gamma_{t,s})}
\left(1-\sum_{n=1}^N\sum_{\tau=1}^t p^{(n,\tau)}_{t,s}w_{(n,\tau),t+d}\right)
\label{cond-2}
\end{eqnarray}
for all $y$.
Therefore, the inequality (\ref{for-1b-1}) is equivalent to the inequality
\begin{eqnarray}
e^{-\eta\lambda(\gamma_{t,s},y)}\ge\sum_{n=1}^N\sum_{\tau=1}^t 
w^{*,s}_{(n,\tau),t}e^{-\eta\lambda(y,c^{(n,\tau)}_{t,s})},
\label{for-1ba}
\end{eqnarray}
where
\begin{eqnarray}\label{for-1bb}
w^{*,s}_{(n,\tau),t}=\frac{p^{(n,\tau)}_{t,s} w_{(n,\tau),t+d}}{\sum_{n'=1}^N
\sum_{\tau'=1}^t p^{(n',\tau')}_{t,s} w_{(n',\tau'),t+d}}.
\end{eqnarray}
According to the aggregating algorithm rule 
we can define $\gamma_{t,s}=\subst(\c_{s,t},\w^{*,s}_t)$ for $1\le s\le d$
such that (\ref{for-1ba}) and its equivalent (\ref{for-1b-1}) are valid.
Here $\subst$ is the substitution function and
$$\w^{*,s}_t=(w^{*,s}_{(n,\tau),t}: 1\le n\le N,1\le\tau\le t),$$ 
$$\c_{t,s}=(c^{(n,\tau)}_{t,s}: 1\le n\le N,1\le\tau\le t).$$ 

The outcomes $y_{t+1},\dots ,y_{t+d}$ will be fully revealed only at 
the time moment $t+d$. The inequality (\ref{for-1b-1}) holds for $y=y_{t+s}$ 
and for the forecasts $\gamma_{t,s}$ for all $1\le s\le d$. By convexity of the exponent 
the inequality (\ref{for-1b-1}) implies that
\begin{eqnarray}
e^{-\eta\lambda(y_{t+s},\gamma_{t,s})}\ge
\sum_{n=1}^N\sum_{\tau=1}^\infty e^{-\eta E_{\p^{(n,\tau)}_{t,s}}
[\lambda(y_{t+s},\tilde c^{(n,\tau)}_{t,s})]}w_{(n,\tau),t+d}.
\label{for-1b-2}
\end{eqnarray} 
holds for all $1\le s\le d$. We use the generalized H\"older inequality and obtain  
\begin{eqnarray}
e^{-\eta\frac{1}{d}\sum\limits_{s=1}^d\lambda(y_{t+s},\gamma_{t,s})}\ge
\sum_{n=1}^N\sum_{\tau=1}^\infty e^{-\eta \frac{1}{d}\sum\limits_{s=1}^d 
E_{\p^{(n,\tau)}_{t,s}}[\lambda(y_{t+s},\tilde c^{(n,\tau)}_{t,s})]}
w_{(n,\tau),t+d}.
\label{for-1b-2a}
\end{eqnarray} 
For more details of the H\"older inequality see \ref{vector-pred-1}.
The inequality (\ref{for-1b-2a}) can be rewritten as
\begin{eqnarray}
e^{-\eta h_{t+d}}\ge
\sum_{n=1}^N\sum_{\tau=1}^\infty e^{-\eta \hat l^{(n,\tau)}_{t,s}}
w_{(n,\tau),t+d},
\label{for-1b-2ab}
\end{eqnarray}
where 
$$
h_{t+d}=\frac{1}{d}\sum_{s=1}^d\lambda(y_{t+s},\gamma_{t,s})
$$
is the (averaged) loss of the aggregating algorithm suffered
on the time interval $[t+1,t+d]$ and
$$
\hat l^{(n,\tau)}_{t,s}=\frac{1}{d}\sum\limits_{s=1}^d 
E_{\p^{(n,\tau)}_{t,s}}[\lambda(y_{t+s},\tilde c^{(n,\tau)}_{t,s})]
$$
is the (averaged) mean loss of the expert $(n,\tau)$.

The protocol of algorithm for aggregating forecasts of experts $(n,\tau)$ is shown below.

\medskip
            
{\bf Algorithm 1}
{\small
\medskip\hrule\hrule\medskip

\noindent Set $w_{(n,\tau),1}=\frac{1}{N}\nu(\tau)$, where $\nu(\tau)=\frac{1}{\tau (\tau+1)}$,
${n=1,\dots , N}$, ${\tau=1,2,\dots}$.

\noindent {\bf FOR} $t=1,\dots ,T$

\hspace{2mm}{\bf IF} $t\le d$ {\bf THEN} put $l^{(n,\tau)}_t=h_t=0$ for all $n$ and $\tau$. 

\hspace{2mm}{\bf ELSE}
\begin{enumerate}
\item Observe the outcomes ${y_{t-d+1},\dots, y_t}$
and predictions ${\gamma_{t-d}=(\gamma_{t-d,1},\dots ,\gamma_{t-d,d})}$ 
of the learner issued at the time moment $t-d$.

\item Compute the loss ${h_t=\frac{1}{d}\sum_{s=1}^d h_{t,s}}$ of the learner on 
the time segment $[t-d+1,t]$, where ${h_{t,s}=\lambda(y_{t-d+s},\gamma_{t-d,s})}$.

\item Compute the losses 
${l^{(n, \tau)}_t=\frac{1}{d}\sum_{s=1}^d l^{(n,\tau)}_{t,s}}$ of the experts ${(n,\tau)}$
for ${1\le n\le N}$, where for ${1\le s\le d}$ we set ${l^{(n, \tau)}_{t,s}=\lambda(y_{t-d+s},c^n_{\tau,t-d-\tau+s})}$ if ${1\le\tau\le t-d}$
and ${l^{(n,\tau)}_{t,s}=\lambda(y_{t-d+s},\gamma_{t-d,s})}$  if $\tau>t-d$.
\end{enumerate}

\hspace{2mm}{\bf ENDIF}

\begin{enumerate}
\setcounter{enumi}{3}
\item Update weights: 
\begin{eqnarray}
w^\mu_{(n,\tau),t}=\frac{w_{(n,\tau),t}e^{-\eta l^{(n,\tau)}_t}}
{\sum_{n'=1}^N\sum_{\tau'=1}^\infty w_{(n',\tau'),t}e^{-\eta l^{(n',\tau')}_t}}
\label{wei-1}
\end{eqnarray}
for $1\le n\le N$, $1\le\tau<\infty$.\footnote{These weights can be computed
efficiently, since the divisor in (\ref{wei-1}) can be represented
\begin{eqnarray}
\sum_{n'=1}^N\sum_{\tau'=1}^\infty w_{(n',\tau'),t}e^{-\eta l^{(n',\tau')}_t}=
\nonumber
\\
\sum_{n'=1}^N\sum_{\tau'=1}^{t-d} w_{(n',\tau'),t}e^{-\eta l^{(n',\tau')}_t}+
e^{-\eta\lambda(y_{t-d+s},\gamma_{t-d,s})}
\left(1-\sum_{n'=1}^N\sum_{\tau'=1}^{t-d} w_{(n',\tau'),t}\right).
\label{up-w-2}
\end{eqnarray}
}

\item Prepare the weights: $w_{(n,\tau),t+d}=w^\mu_{(n,\tau),t}$ for
$1\le n\le N$ and $1\le\tau<\infty$.


\item Receive predictions $\c^n_\tau$ issued by the experts $1\le n\le N$ at the 
time moments $\tau\le t$ and their confidence levels $\p^n_\tau$ . 

\item Extract the segments of forecasts 
$\c^{(n,\tau)}_t=(c^{(n,\tau)}_{t,1},\dots ,c^{(n, \tau)}_{t,d})$ of the the auxiliary 
experts $(n,\tau)$, where $c^{(n,\tau)}_{t,s}=c^n_{\tau,t-\tau+s}$ for $1\le\tau\le t$,
and the segments of the corresponding confidences 
$\p^{(n,\tau)}_t=(p^{(n,\tau)}_{t,1},\dots , p^{(n,\tau)}_{t,d})$,
where $p^{(n,\tau)}_{t,s}=p^n_{\tau,t-\tau+s}$.\footnote
{Here $c^n_{\tau,t-\tau+s}$ is a forecast of
the real expert $n$ for the time moment $t+s$ issued at the time moment $\tau$ 
and $1\le s\le d$.} 

\item Compute long-term forecast 
$\gamma_t=(\gamma_{t,1},\dots ,\gamma_{t,d})$ of the learner, where
\begin{eqnarray}
\gamma_{t,s}=\subst(\c_{s,t},\w^{*,s}_{t,s}),
\nonumber
\\
\w^*_{t,s}=(w^{*,s}_{(n,\tau),t}: 1\le n\le N,1\le\tau\le t),
\nonumber
\\
w^{*}_{(n,\tau),t}=\frac{p^{(n,\tau)}_{t,s} w_{(n,\tau),t+d}}{\sum_{n'=1}^N
\sum_{\tau'=1}^t p^{(n',\tau')}_{t,s} w_{(n',\tau'),t+d}},
\label{w-u-3}
\\
{\c_{t,s}=(c^{(n,\tau)}_{t,s}: 1\le n\le N,1\le\tau\le t)}
\nonumber
\end{eqnarray} 
for $1\le s\le d$.\footnote{
For computation the values of the function $\subst$, we can use the rules 
(\ref{subst-1a}) or~(\ref{subst-2a}) from Section \ref{prel-1}.
}

\end{enumerate}
\noindent \hspace{2mm}
{\bf ENDFOR}
\medskip\hrule\hrule\medskip
}

Denote for $t>d$
$$
l^{(n,\tau)}_{t,s}=\lambda(y_{t-d+s},c^{(n,\tau)}_{t-d,s}),
$$
$$
\tilde l^{(n,\tau)}_{t,s}=\lambda(y_{t-d+s},\tilde c^{(n,\tau)}_{t-d,s}),
$$
$$
\hat l^{(n,\tau)}_{t,s}=E_{\p^{(n,\tau)}_{t-d,s}}[\tilde l^{(n,\tau)}_{t,s}],
$$
$$
h_{t,s}=\lambda(y_{t-d+s},\gamma_{t-d,s}).
$$
We put these quantities to be 0 for $1\le t\le d$.
Also, $\hat l^{(n,\tau)}_{t,s}=l^{(n,\tau)}_{t,s}=h_{t,s}$ for ${\tau>t}$.
Since by definition 
$$\hat l^{(n,\tau)}_{t,s}=
p^{(n,\tau)}_{t-d,s}l^{(n,\tau)}_{t,s}+(1-p^{(n,\tau)}_{t-d,s})h_{t,s},$$
we have $$h_{t,s}-\hat l^{(n,\tau)}_{t,s}=p^{(n,\tau)}_{t-d,s}(h_{t,s}-l^{(n,\tau)}_{t,s}).$$

Recall that $h_t=\frac{1}{d}\sum\limits_{s=1}^d h_{t,s}$ be the algorithm (average) 
loss and 
$\hat l^{(n,\tau)}_t=\frac{1}{d}\sum\limits_{s=1}^d \hat l^{(n,\tau)}_{t,s}$ be 
the (average) loss of the auxiliary expert $(n,\tau)$.

Define the discounted (average) excess loss with respect to an expert $(n,\tau)$ 
at a time moment $t>d$ by
\begin{eqnarray}
r^{(n,\tau)}_t=h_t-\hat l^{(n,\tau)}_t.
\label{excess-1}
\end{eqnarray}
By definition of $\hat l^{(n,\tau)}_t$ we can represent the discounted 
excess loss (\ref{excess-1}) as
\begin{eqnarray*}
r^{(n,\tau)}_t=\frac{1}{d}\sum\limits_{s=1}^d
p^{(n,\tau)}_{t-d,s}(h_{t,s}-l^{(n,\tau)}_{t,s})=
\nonumber
\\
\frac{1}{d}\sum\limits_{s=1}^d p^{(n,\tau)}_{t-d,s}(\lambda(y_{t-d+s},\gamma_{t-d,s})-
\lambda(y_{t-d+s},c^{(n,\tau)}_{t-d,s})).
\end{eqnarray*}
We measure the performance of our algorithm by the cumulative discounted (average) excess 
loss with respect to any expert $(n,\tau)$.
\begin{theorem}\label{main-3fa}
For any ${T\ge d+1}$ and ${1\le n\le N}$, the following upper bound 
for the cumulative excess loss holds true:
\begin{eqnarray}
\sup_{\tau\le T-d}\sum\limits_{t=\tau+d}^T r^{(n,\tau)}_t\le\frac{d}{\eta}\left(\ln N+2\ln (T-d+1)\right).
\label{mTT-1afaa}
\end{eqnarray}
\end{theorem}
{\it Proof}.
Let $m_t=-\frac{1}{\eta}\ln\sum_{n=1}^N\sum_{\tau=1}^{T-d}
w_{(n,\tau),t}e^{-\eta\hat l^{(n,\tau)}_t}$. 
Let us apply Corollary \ref{prop-1} from 
Section \ref{vector-pred-1} for the case where $(n,\tau)$ are experts for 
$1\le n\le N$ and $1\le\tau\le T-d$, so, $M=N(T-d)$. 
Also, set 
$$
\l_t=\hat\l_t=(\hat l^{(n,\tau)}_t: 1\le n\le N, 1\le\tau\le T-d)
$$ 
and $\q$ be unit vector
of length $N(T-d)$ whose $(n,\tau)$th coordinate is 1. 
By (\ref{for-1b-2ab}) $h_t\le m_t$. Then by (\ref{iineq-1})
\begin{eqnarray}
\sum\limits_{t=1}^T h_t-\sum\limits_{t=1}^T\hat l^{(n,\tau)}_t\le
\frac{d}{\eta}\ln (N(T-d)(T-d+1))
\label{inn-1}
\end{eqnarray}
for each expert $(n,\tau)$ such that $1\le n\le N$ and $\tau\le T-d$.

Since $\hat l^{(n,\tau)}_t=h_t$ for $t<\tau+d$, using (\ref{excess-1}),
we obtain (\ref{mTT-1afaa}).
$\triangle$

\section{Online Smoothing Regression}\label{regr-1}

In this section we consider the online learning scenario within the supervised setting
(that is, data are pairs $(x,y)$ of predictor-response variables). 
A forecaster presents a regression function $F$ defined
on a set $X$ of objects, which are called signals. After a pair $(x,y)$
be revealed the forecaster suffers a loss $\lambda(y,F(x))$, 
where $\lambda(y,\gamma)$ is some loss function. We assume that $y\in {\cal R}$ and 
that the loss function is $\eta$-mixable for some $\eta>0$.

An example is a linear regression, where $X\subseteq {\cal R}^k$ is a set of 
$k$-dimensional vectors and a regression function is
a linear function $F(\x)=(\w\cdot\x)$, where $\w\in {\cal R}^k$ is a weight vector
and $\lambda(y,F(\x))=((\w\cdot\x)-y)^2$ is the square loss.

In the online mode, at any step $t$, to define the forecast for step $t+1$ -- a regression function $F_{t+1}(x)$, we use the prediction with expert advice approach. A feature of this approach is that 
we aggregate the regression functions $F^1_\tau (x)$ for $1<\tau\le t+1$, each of 
which depends on the segment $(x_1,y_1),\dots ,(x_{\tau-1},y_{\tau-1})$ of the sample.
At the end of step $t$ we define (initialize) the next regression function $F^1_{t+1}(x)$
by the sample $(x_1,y_1),\dots ,(x_t,y_t)$.

Since the forecast $F_{t+1}(x)$ can potentially be applied to any future input value $x$, 
we consider this method as a kind of long-term forecasting.

We briefly describe below the changes made in Algorithm 1.
We introduce signals in the protocol from Section \ref{v-s-1}.

\medskip
            
{\bf Algorithm 2}
{\small
\medskip\hrule\hrule\medskip

Set initial weights $w_{\tau,1}$ as in Algorithm 1.

\noindent {\bf FOR} $t=1,\dots ,T$


\begin{enumerate}
\item
Observe the pair $(x_t,y_t)$ 
and compute the losses suffered by the learner 
$h_t=\lambda(y_t,F_t(x_t))$ and by the expert regression functions: $l^\tau_t=\lambda(y_t,F^1_\tau(x_t))$ 
if $1\le\tau\le t$ and $l^{\tau}_t=\lambda(y_t,F_t(x_t))$ otherwise.


\item
Update weights: 
\begin{eqnarray}
w_{\tau,t+1}=w^\mu_{\tau,t}=\frac{w_{\tau,t}e^{-\eta l^{\tau}_t}}
{\sum_{\tau'=1}^\infty w_{\tau',t}e^{-\eta l^{\tau'}_t}}
\label{wei-1a}
\end{eqnarray}
for $1\le\tau<\infty$. See also footnote to (\ref{wei-1}). 

\item
Initialize the next regression function $F^1_{t+1}(x)$ using the sample $(x_1,y_1),\dots ,(x_t,y_t)$ and
define the forecast of the learner for step $t+1$ 
\begin{eqnarray}
F_{t+1}(x)=\subst(\F_t(x),\w^*_t) \mbox{ for any } x\in X, 
\label{gen-for-1a}
\end{eqnarray}
where $\F_t(x)=(F^1_\tau (x): 1\le\tau\le t+1)$,
$\w^*_t=(w^*_{\tau,t+1}: 1\le\tau\le t+1)$, and 
$w^*_{\tau,t+1}=\frac{w_{\tau,t+1}}{\sum_{\tau'=1}^{t+1} w_{\tau',t+1}}$
for $1\le\tau\le t+1$.\footnote{We extend the rules (\ref{subst-1a}) and (\ref{subst-2a}) to
functional forecasts in a natural way, see (\ref{subst-1abc}) below. See also, the footnote for item 8 of Algorithm 1}
\end{enumerate}

\noindent \hspace{2mm}

{\bf ENDFOR}
\medskip\hrule\hrule\medskip
}

For the square loss $\lambda(y,\gamma)=(y-\gamma)^2$, where $y\in [-B,B]$, by (\ref{subst-1a})
the regression function (\ref{gen-for-1a}) can be defined in the closed form:
\begin{eqnarray}
F_{t+1}(x)=
\frac{1}{4\eta B}\ln\frac{\sum\limits_{\tau=1}^{t+1} w_{\tau,t+1} 
e^{-\eta (B-F^1_\tau(x))^2}}
{\sum\limits_{\tau=1}^{t+1} w_{\tau,t+1} e^{-\eta (B+F^1_\tau(x))^2}}
\label{subst-1abc}
\end{eqnarray}
for each $x$ or by the rule (\ref{subst-2a}).\footnote{The most appropriate choices of $\eta$ are $\eta=\frac{1}{2B^2}$
for the rule (\ref{subst-1a}) and $\eta=\frac{1}{8B^2}$ for~(\ref{subst-2a}).
The more straightforward definition (\ref{subst-2a}) results in four times more regret 
but easier for computation.}

Let us analyze the performance of Algorithm 2 as a forecaster on $d$ steps ahead.

For any time moment $t\ge d$ a sequence $(x_{t-d+1},x_{t-d+1}),\dots ,(x_t,y_t)$ is revealed.
Denote by $h_t=\frac{1}{d}\sum\limits_{s=1}^d\lambda(y_{t-s+1},F_\tau(x_{t-s+1}))$ 
the average loss of the learner on time interval $[t-d+1,t]$ and by 
$l^\tau_t=\frac{1}{d}\sum\limits_{s=1}^d\lambda(y_{t-s+1},F^1_\tau(x_{t-s+1}))$ 
the average loss of any auxiliary expert $\tau\le t$.

The regret bound of Algorithm 2 does not depend on $d$:
\begin{theorem}\label{regression-th-1}
For any $T$,
\begin{eqnarray}
\sup_{1\le\tau\le T-1}\sum\limits_{t=\tau+1}^T h_t-l^{\tau}_t\le\frac{2}{\eta}\ln T.
\label{mTT-1afaas}
\end{eqnarray}
\end{theorem}
{\it Proof.} The analysis of the performance of Algorithm 2 for the case of prediction on $d$ steps 
ahead is similar to that of Algorithm 1 for $d=1$. 
Let $T$ and $d\le\tau<T$ be given.
Using the technics of Section \ref{vector-pred-1}, we obtain for any $1\le s\le d$,
$$
\sum\limits_{t=\tau+1}^T \lambda(y_{t-s+1},F_\tau(x_{t-s+1}))-
\lambda(y_{t-s+1},F^1_\tau(x_{t-s+1}))\le\frac{1}{\eta}\ln (T(T-1)).
$$
Summing this inequality by $s=1,\dots ,d$ and dividing by $d$, we obtain (\ref{mTT-1afaas}).
$\triangle$

In particular, Theorem \ref{regression-th-1} implies that the total loss of Algorithm 2 
at any time interval $[1,T]$ is no more (up to logarithmic regret) than the loss of 
the best regression algorithm constructed in the past. 

{\bf Online regression with a sliding window.}
Some time series show a strong dependence on the latest information instead of all the data. 
In this case, it is useful to apply regression with a sliding window.
In this regard, we consider the application of Algorithm 2 for the case 
of online regression with a sliding window. The corresponding expert represents some type of
dependence between input and output data. If this relationship is relatively regular
the corresponding experts based on past data can successfully compete with experts based on 
the latest data. Therefore, it may be useful to aggregate the predictions of all the auxiliary 
experts based on past data.

Let $F^1_t(\x)=(\w_t\cdot\x)$ be the ridge regression function, where for $t>h$,
$\w_t=\left(\sigma I+X'_t X_t\right)^{-1} X_t'\y_t$. Here $X_t$ is the 
matrix in which rows are formed by vectors $\x_{t-h},\dots , \x_{t-1}\in {R}^k$ 
($X'_t$ is the transposed matrix $X_t$), $I$ is a unit matrix, $\sigma>0$ is a parameter, 
and $\y_t=(y_{t-h},\dots , y_{t-1})$. For $t\le h$ we set $\w_t$ equal to some 
fixed value.

We use the square loss function and assume that $y_t\in [-B,B]$ for all $t$.
For each $t$ we define the aggregating regression function $F_{t+1}$ 
(the learner forecast) by (\ref{subst-1abc}) using the regression functions 
$F^1_\tau$ (the expert strategies) for $h<\tau\le t+1$, where each such a function
is defined using a learning sample (a window) 
$(\x_{\tau-h},y_{\tau-h}),\dots , (\x_{\tau-1},y_{\tau-1})$.\footnote{
The computationally efficient algorithm for recalculating matrices during 
the transition from $X_t$ to $X_{t+1}$ for some special type of online regression with 
a sliding window was presented by \cite{arce2012online}. Similar effective options for regression 
using Algorithm 2 can also be developed.
}

{\bf Experiments.}
Let us present the results of experiments which were performed on synthetic data.
The initial data was obtained as a result of sampling from a data generative
model.

We start from a sequence $\x_1,\dots ,\x_T$ of $20$-dimensional signals sampled i.i.d
from the multidimensional normal distribution. The signals are revealed online and $T=3000$.

The target variable $y$ is generated as follows. First, three random linear dependencies 
are generated, i.e. three weights vectors $\w_1,\w_2,\w_3$ are generated
(so $y_t=(\w_\tau\cdot \x_t)$ for $\tau=1,2,3$ on the corresponding time intervals).
The time scale $[1,T]$ is divided into $K=7$ random consecutive parts. On each interval data 
is generated based on one of these three random regressions $y_t=(w_\tau\cdot x_t)+\epsilon$, 
where $\tau=1,2$ or $3$ and $\epsilon$ is a low noise. 
That is, the dependence of $y$ on $x$ is switched $7$ times.

Each expert $F^1_\tau(\x)$ corresponds to a linear regression trained in a sliding data window
$(\x_ {\tau-h}, y_ {\tau-h}), ..., (\x_\tau, y_\tau)$ of length $h=40$. 
There are a total of $T-h+1$ experts.

\begin{figure}[!htb]
\includegraphics[width=0.99\textwidth]{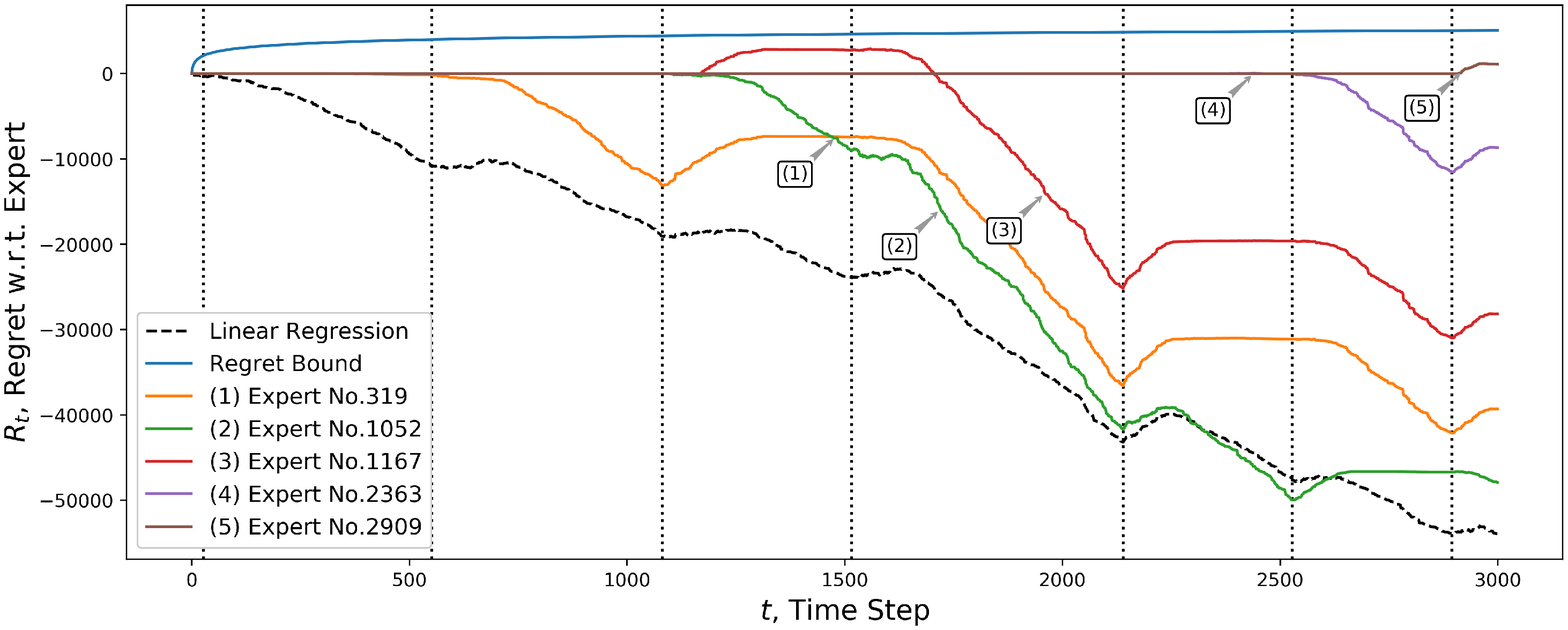}

        \caption{{\small
The graphs of the regret $R^\tau_t=H_t-L^\tau_t$ for the experts
starting at $\tau=319,1052,1167,2363,2909$
time moments. 
The theoretical upper bound for the regret is represented by the line located 
above all the lines in the graph. The regret with respect to the simple linear 
regression is presented by the dotted line
}}\label{fig-1}
                  \end{figure}

Figure \ref{fig-1} shows the results of the random experiment,
where the graphs of $H_t-L^\tau_t$ present the regret of Algorithm~2 with respect to the experts
starting at several time moments $\tau$. 

The regret with respect to the simple linear regression performed on all data
interval is also presented. We see that Algorithm~2 efficiently adapts to data and also outperforms linear regression on the entire dataset.
The theoretical upper bound for the regret is also plotted 
(it is clear that all lines are below it).

\section{Conclusion}

In the paper we have developed the aggregating algorithm for long-term interval forecasting 
which is capable of combining current predictions of the experts with the outdated 
ones (made earlier). Combining past and current long-term forecasts allows to protects 
the algorithm from temporary changes in the trend of the time series, noise and outliers. 
Our mechanism can be applied to the time series forecasting models that are capable 
of predicting for the infinitely many time moments ahead, e.g. widespread ARMA-like models.
For the developed algorithm we proved the sublinear $O(\ln T)$ regret bound.

We have applied PEA approach for the case of online supervised learning,
where instead of point predictions, the experts and the learner present predictions 
in the form of regression functions. The method for smoothing regression using 
expert advice was presented. We consider this method of regression as a kind of 
long-term forecasting. 
Experiments conducted on synthetic data show the effectiveness of the proposed method.

\section*{Acknowledgments}

\noindent The research was partially supported by the Russian Foundation for Basic Research grant 16-29-09649 ofi m.



\bibliographystyle{elsarticle-num-names}


\appendix

\section{Auxiliary results}\label{vector-pred-1}

{\bf Vector-valued forecasts}.
In this paper we aggregate the vector forecasts. To do this, following \cite{Kaln2017}, 
we apply the aggregation rule to each coordinate separately. Since the loss function is 
$\eta$-mixable, for any time moment $t$ for each $1\le s\le d$ a prediction 
$\gamma_{t,s}$ exists such that the inequality (\ref{cond-1}) is valid.

Let $y_{t+1},\dots ,y_{t+d}$ be a sequence of outcomes.
Multiplying the inequalities (\ref{for-1b-2}) for $s=1,\dots , d$ we obtain
\begin{eqnarray}
e^{-\eta\sum_{s=1}^d\lambda(y_{t+s},\gamma_{t,s})}\ge
\prod_{s=1}^d\sum_{n=1}^N\sum_{\tau=1}^\infty e^{-\eta E_{\p^{(n,\tau)}_{t,s}}
[\lambda(y_{t+s},\tilde c^{(n,\tau)}_{t,s})]}w_{(n,\tau),t+d}.
\label{aggr-rule-1fuapp}
\end{eqnarray}

The generalized H\"older inequality says that
$$
\|f_1f_2\cdots f_d\|_r\le\|f_1\|_{q_1}\|f_2\|_{q_2}\cdots\|f_d\|_{q_d},
$$
where $\frac{1}{q_1}+\dots +\frac{1}{q_d}=\frac{1}{r}$,  $q_s\in (0,+\infty)$
and $f_s\in L^{q_s}$ for $1\le s\le d$. Let $q_s=1$ for all $1\le s\le d$, then $r=1/d$.
Let 
$$
f_s=e^{-\eta E_{\p^{(n,\tau)}_{t,s}}
[\lambda(y_{t+s},\tilde c^{(n,\tau)}_{t,s})]}
$$ 
for $s=1,\dots ,d$ and $\|f\|_1=E_{\w}(f)$, where 
$$\w=(w_{(n,\tau),t+d}: 1\le n\le N, \tau\ge 1).$$ 
Then by H\"older inequality we obtain (\ref{for-1b-2a}).

{\bf Regret analysis}.
We use relative entropy as the basic tool for the regret analysis. Let 
$D(\p\|\q)=\sum\limits_{i=1}^N p_i\ln\frac{p_i}{q_i}$
be the relative entropy, where $\p=(p_1,\dots ,p_N)$ and $\q=(q_1,\dots ,q_N)$
are probability vectors.\footnote{That is, $p_i\ge 0$, $q_i\ge 0$ for all $i$
and $\sum\limits_{i=1}^N p_i=1$ and $\sum\limits_{i=1}^N q_i=1$.} We also define $0\ln 0=0$. 

Let $\l_t=(l^1_t,\dots ,l^N_t)$ be the losses of the experts at step $t$, where $1\le i\le N$
and $l^i_t=0$ for $1\le t\le d$. 
Let also, the evolution of the weights $\w^\mu_t$ of the experts be defined
by the rule (\ref{weight-update-1}) and ${\w_{t+d}=\w^\mu_t}$. Denote the dot product of two 
vectors by $(\q\cdot\l_t)$. Let also, 
$m_t=-\frac{1}{\eta}\ln\sum\limits_{i=1}^N w_{i,t} e^{-\eta l^i_t}$ for $t>d$.
\begin{lemma}\label{Kul-L-ineq-3} (see \cite{BoW2002})
For any $t>d$, and for a comparison probability vector $\q$,
\begin{eqnarray}
m_t=(\q\cdot\l_t)+\frac{1}{\eta}(D(\q\|\w_t)-D(\q\|\w^\mu_t)).
\label{MPP-bound-1}
\end{eqnarray}
\end{lemma}
{\it Proof}.
By method (\ref{weight-update-1}) of the weights updating,
the equality (\ref{MPP-bound-1}) is obtained as follows:
\begin{eqnarray*}
m_t-\sum\limits_{i=1}^N q_i l^i_t=\sum\limits_{i=1}^N q_i
\left(\frac{1}{\eta}\ln e^{-\eta l^i_t}+m_t\right)=
\\
\frac{1}{\eta}\sum\limits_{i=1}^N q_i\left(\ln e^{-\eta l^i_t}-
\ln\sum\limits_{j=1}^N w_{j,t}e^{-\eta l^j_t}\right)=
\\
\frac{1}{\eta}\sum\limits_{i=1}^N q_i\ln\frac{e^{-\eta l^i_t}}
{\sum\limits_{j=1}^N w_{j,t}e^{-\eta l^j_t}}=
\\
\frac{1}{\eta}\sum\limits_{i=1}^N q_i\ln\frac{w^\mu_{i,t}}{w_{i,t}} =
\frac{1}{\eta}(D(\q\|w_t)-D(\q\|w^\mu_t)).
\end{eqnarray*}

\begin{corollary}\label{prop-1} Let $T$ be a forecasting horizon and $N$ be a number 
of the experts. Consider the case of $d$-steps ahead prediction for $d\ge 1$.
Then for any comparison vector $\q$,
\begin{eqnarray}
\sum\limits_{t=1}^T m_t-\sum\limits_{t=1}^T (\q\cdot\l_t)\le\frac{1}{\eta}d
\max_{1\le i\le N, 1\le t\le d}\ln\frac{1}{w^\mu_{i,t}}.
\label{iineq-1}
\end{eqnarray}
\end{corollary}
{\it Proof}.
Summing (\ref{MPP-bound-1}) for $d+1\le t\le T$, we obtain
\begin{eqnarray}
\sum\limits_{t=d+1}^T m_t-\sum\limits_{t=d+1}^T (\q\cdot\l_t)=
\nonumber
\\
\frac{1}{\eta}\sum\limits_{t=d+1}^T(D(\q\|\w_t)-D(\q\|\w^\mu_t))=
\label{enequa-2}
\\
\frac{1}{\eta}\sum\limits_{t=d+1}^T(D(\q\|\w^\mu_{t-d})-D(\q\|\w^\mu_t))=
\label{enequa-3}
\\
\frac{1}{\eta}\sum\limits_{t=1}^d D(\q\|\w^\mu_t)\le
\frac{1}{\eta}d\max_{1\le j\le N, 1\le t\le d}\ln\frac{1}{w^\mu_{j,t}}.
\label{enequa-4}
\end{eqnarray}
In transition from (\ref{enequa-2}) to (\ref{enequa-3}) we use equality
$\w_t=\w^\mu_{t-d}$ for $d<t\le T$. In transition from (\ref{enequa-3}) to (\ref{enequa-4}) 
the positive and corresponding negative terms telescope and only first $d$ positive
terms remain. Also, we have used the inequality
\begin{eqnarray}
D(\q\|\p)=\sum\limits_{i=1}^N q_i\ln\frac{q_i}{p_i}=
\sum\limits_{i=1}^N q_i\ln q_i+\sum\limits_{i=1}^N q_i\ln\frac{1}{p_i}\le
\max_{1\le i\le N}\ln\frac{1}{p_i} 
\nonumber
\end{eqnarray}
for all probability vectors $\q$ and $\p$.

Since $m_t=l^i_t=0$ for all $1\le t\le d$ and $1\le i\le N$, (\ref{iineq-1}) follows.
$\triangle$

\end{document}